\documentclass{article}

\usepackage[preprint]{neurips_2024}

\usepackage[utf8]{inputenc} 
\usepackage[T1]{fontenc}    
\usepackage{url}            
\usepackage{booktabs}       
\usepackage{amsfonts}       
\usepackage{nicefrac}       
\usepackage{xcolor}         
\usepackage{graphicx}
\usepackage{subcaption}
\usepackage{array}
\usepackage{kotex}
\usepackage{amsmath,amssymb,amsfonts}
\usepackage{algorithmic}
\usepackage{hyperref}       
\setcitestyle{numbers}
\title{A Data-Driven Odyssey in Solar Vehicles}

\author{
  Do Young Kim\textsuperscript{}\thanks{These authors contributed equally to this work.}\\
  \texttt{doyoung42@snu.ac.kr}\\
  Seoul National University,\\
  Seoul, Korea
  \And
  Kyunghyun Kim\textsuperscript{}\footnotemark[1]\\
  \texttt{khkimamy@snu.ac.kr}\\
  Seoul National University, \\
  Seoul, Korea
  \And
  Gyeongseop Lee\textsuperscript{}\\
  \texttt{pepsipower@snu.ac.kr}\\
  Seoul National University,\\
  Seoul, Korea
  \And
  Niloy Das\textsuperscript{}\\
  \texttt{dasn1@unlv.nevada.edu}\\
  University of Nevada,\\
  Las Vegas, USA
  \And
  Seong-Woo Kim\textsuperscript{}\thanks{Corresponding author.}\\
  \texttt{snwoo@snu.ac.kr}\\
  Seoul National University, \\
  Seoul,  Korea
}

\begin{document}

\maketitle

\begin{abstract}
Solar vehicles, which simultaneously produce and consume energy, require meticulous energy management. However, potential users often feel uncertain about their operation compared to conventional vehicles. This study presents a simulator designed to help users understand long-distance travel in solar vehicles and recognize the importance of proper energy management. By utilizing Google Maps data and weather information, the simulator replicates real-world driving conditions and provides a dashboard displaying vehicle status, updated hourly based on user-inputted speed. Users can explore various speed policy scenarios and receive recommendations for optimal driving strategies. The simulator's effectiveness was validated using the route of the World Solar Challenge (WSC). This research enables users to monitor energy dynamics before a journey, enhancing their understanding of energy management and informing appropriate speed decisions.

\end{abstract}

\section{Introduction}
The transition from internal combustion engine vehicles to electric vehicles has heightened interest in the management of energy storage and control systems. As if the transition has introduced challenge issues \cite{marzbani2023electric}\cite{kim2011design}, solar vehicles introduce unique challenges. Research on conventional vehicles has focused on minimizing stored energy use \cite{choi2011energy}, and planning routes based on fuel or charging stations \cite{zhou2022location}. In contrast, solar vehicles generate and consume energy simultaneously during driving, necessitating stable speed policies that consider energy production, consumption, and battery status for long-distance travel. Energy production varies greatly with weather conditions, and consumption changes with the driving environment. However, as noted in \cite{kaplan2016time}, little research has been conducted on energy harvesting and power management for long-term solar vehicle driving, which makes it difficult to verify energy changes and reducing driving reliability.

\begin{figure}
    \centering
    \includegraphics[width=\columnwidth]{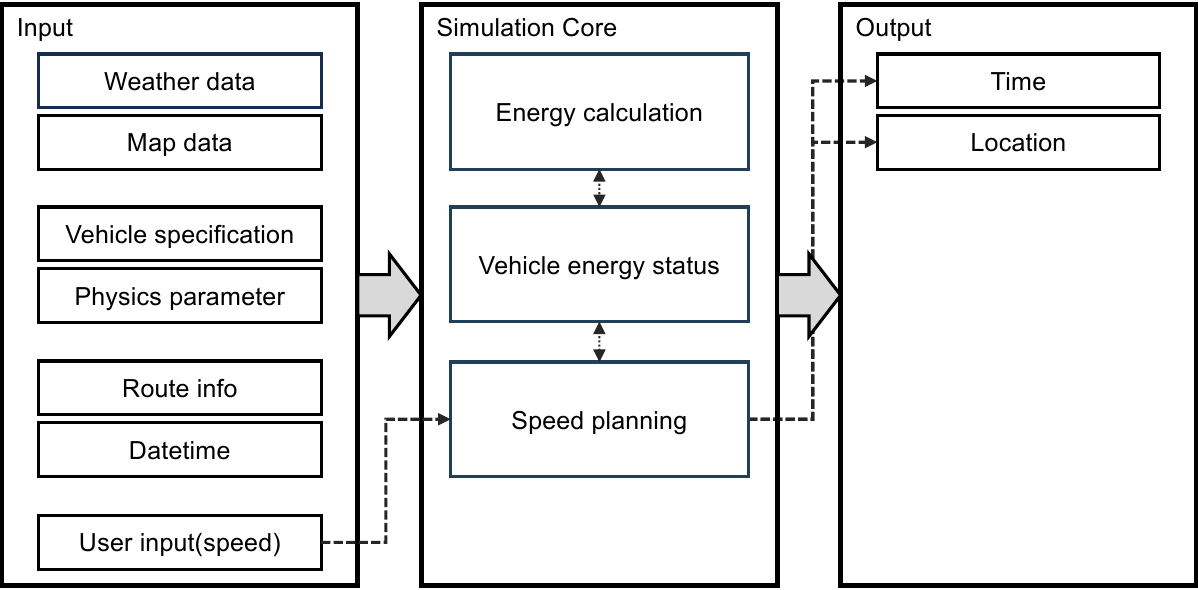}
    \caption{Block diagram for the simulator's data processing flow.}
 \label{fig1}
\end{figure}

To address these issues, we developed a simulator that enables users to indirectly experience driving a solar vehicle and make informed speed decisions. As shown in Figure \ref{fig1}, the simulator calculates expected energy production and consumption based on user-inputted speed, utilizing data on weather, maps, and vehicle specifications for the intended route, and updates the vehicle's status over time. This allows users to verify computed speed policies and observe driving results for their chosen speeds. We validated the simulator using the route of the World Solar Challenge (WSC) in Australia, enabling prior verification of speed policy strategies and emphasizing the importance of energy management in solar vehicles.

\section{Related works}


\begin{figure} [ht]
    \centering
    \includegraphics[scale=0.75]{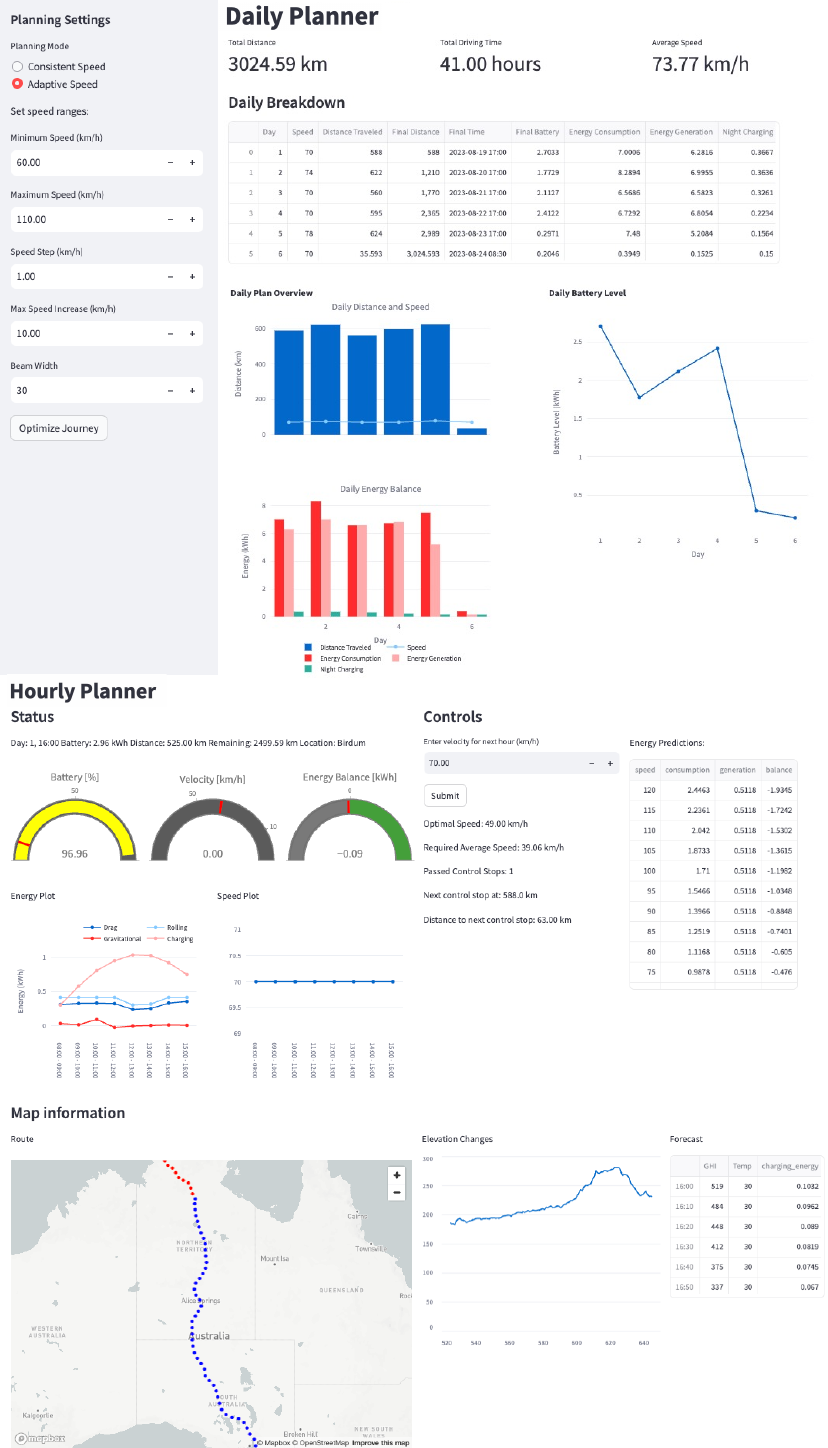}
    \caption{Interface of the solar vehicle race simulator using the WSC route.}
\label{fig:whole}
\end{figure}

Research on solar vehicles has been ongoing to establish sustainable mobility. Unlike conventional vehicles, solar vehicles simultaneously produce and consume energy, leading to varying driving ranges based on energy management and vehicle components \cite{karoui2023integrated}. The distance a solar vehicle can travel may vary by up to 2.5 times depending on climate conditions \cite{yamaguchi2021analysis}, and solar energy production can differ by up to 2.2 times due to temperature variations \cite{shaker2024examining}. While many studies have modeled energy production and consumption using physical equations, managing energy consumption considering variable production remains challenging. To address this, methods for predicting energy production or detecting changes in solar irradiance have been proposed, along with energy management strategies based on these predictions \cite{martins2022systematic}\cite{park2021prediction}.
However, previous studies have often focused solely on either energy production or consumption, or presented driving policies with specific objectives, limiting a comprehensive understanding of energy variations in solar vehicles.

Solar vehicle technologies have been demonstrated through solar driving competitions. Simulators that reflect driving conditions and consider energy production and consumption have been developed, proposing optimal driving policies. Yet, they are limited in responding to diverse situations, and policy changes during driving are not easily accommodated \cite{lv2016speed}\cite{liu2014power}. Consequently, drivers may struggle with decision-making in unplanned scenarios. If drivers could experience various driving conditions beforehand, they could make rational decisions without hesitation. Additionally, a supporter trained in diverse scenarios could aid in producing positive outcomes.
We considered that a simulator with a human interface, where solar vehicle drivers can directly input speed with real-world conditions, would enable users to learn indirectly. Through the data generated from the simulator,  we expect to derive the foundation for autonomous solar vehicles capable of optimized driving in various situations.

\section{Architecture of the simulator}
To enable drivers and prospective users to empirically experience the unknown characteristics of the vehicle, we developed a simulator based on real environmental data. The simulator is implemented using Streamlit\cite{streamlit}, a web-based platform. Streamlit is a rapid prototyping tool that uses Python to freely design UX, allowing for the separation of data processing, and user interface, thereby enhancing maintainability and scalability. This study aims to enable solar vehicle drivers to increase their understanding of energy production and consumption and to indirectly experience driving. The simulator architecture is organized into the following stages.

\subsection{Dataset structure}
We utilize map data and weather data. Map data includes latitude, longitude, altitude, area names, and weather-corresponding area names of route nodes. This information, retrieved via the Google Maps API~\cite{googlemaps}, provides essential details for estimating the energy required for driving, such as route length, slope, and curvature. Weather data from Solcast~\cite{solcast} comprises atmospheric temperature, Global Horizontal Irradiance (GHI), wind direction, and wind speed, matched to each route node to calculate accurate energy production and consumption based on time and location.

\subsection{Simulator configuration}

The simulator consists of three stages: 1) Setting vehicle specifications and physical parameters, 2) Establishing a driving plan, and 3) Outputting and controlling driving information.

\subsubsection{Vehicle specifications and physical parameters}
Users can set the vehicle's specifications and physical parameters in the simulator.
Users input information such as the area and efficiency of the solar panels, vehicle weight, air resistance coefficient, rolling resistance coefficient, and battery capacity, which affect the vehicle's energy production and consumption, allowing them to apply various vehicle designs. Since solar vehicles are not yet commercially available and are custom-built, their appearances and specifications vary. By adjusting specifications, users can observe the effects of vehicle weight and solar cells, verifying how different designs affect driving conditions and providing a basis for design insights.

\subsubsection{Establishing a driving plan}
\begin{figure}[ht]

\centering
\includegraphics[scale=0.7]{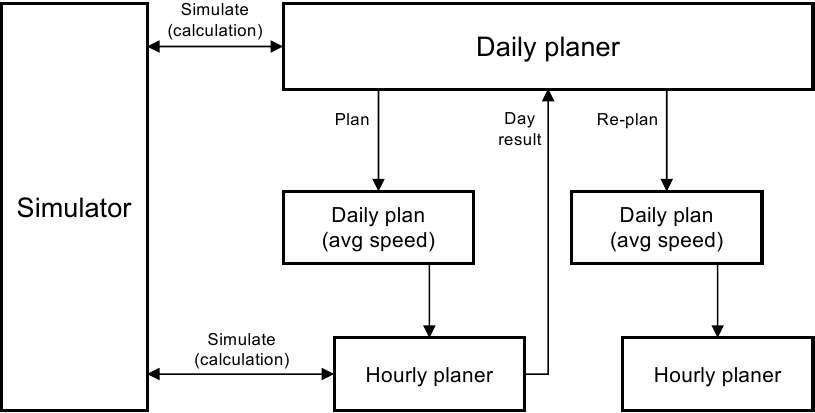}

\caption{Planner that adjusts long-term and daily schedules.}
\label{fig:planner}
\end{figure}

The driving plan is created in two stages: a daily average speed plan and, based on it, an hourly speed plan. The simulator uses Beam Search \cite{ow1988filtered} to pre-establish the daily average speed for the route, giving users guidance for solar vehicle driving. Subsequently, users can then set daily driving goals according to the suggested speeds, as shown on the left side of Figure\ref{fig:whole}. Speed input ranges are limited, with increments of 1 km/h. After completing the daily driving plan, re-planning is possible. In the configuration shown in Figure \ref{fig:planner}, users can adjust the plan for the remaining route based on simulation results after each day's drive.

\subsubsection{Driving information display and control}

On the driving information screen (Figure \ref{fig:whole}), users can monitor and control the vehicle’s status in real time. It shows battery status, driving speed, and energy balance, allowing users to check after each 1-hour segment. Graph display factors like air resistance, rolling resistance, and gravitational resistance to illustrate energy consumption. Users can input driving speeds for each segment and adjust the vehicle's energy state, monitoring the remaining distance and total distance driven. The simulator provides information on predicted energy changes and GHI forecasts to help users choose the optimal driving speed. Additionally, the map displays the vehicle's location and route, driven sections and altitude changes to highlight how terrain affects energy consumption.

\subsection{Energy status}
In this section, we explain the energy calculation formulas applied in the simulator, reviewing the logic and making necessary corrections. The equations are formulated based on the works by Thacher~\cite{thacher2015solar}, Carrol~\cite{carrol2003winning}, and Guerrero and Duarte-Mermoud~\cite{guerrero2016online}. 

\paragraph{Energy calculations of a solar vehicle hourly base}
To illustrate the energy changes in a solar vehicle, it is necessary to calculate the energy generated and consumed on an hourly and daily basis. The total energy of the solar vehicle is the sum of its hourly energy values:
\begin{align}
&E_{\text{charge}} = \sum_{i} E_{\text{charge}}(r_i, t_i),\\
&E_{\text{consumption}} = \sum_{i} E_{\text{consumption}}(r_i, t_i),
\end{align}
where $E_{\text{charge}}(r_i, t_i)$ and $E_{\text{consumption}}(r_i, t_i)$ represent the hourly energy charged and consumed, respectively, at route $r_i$ and time $t_i$ during operating time $i$.

The energy generated at a time is:
\begin{equation}
E_{\text{charge}}(r_i, t_i) = \eta_{\text{system}} \eta_{\text{panel}}\text{GHI}(r_i, t_i) A_{\text{panel}},\end{equation}
where $\eta_{\text{system}}$ is overall system efficiency, $\eta_{\text{panel}}$ is panel efficiency, \text{GHI} is  Global Horizontal Irradiation and $A_{\text{panel}}$ is panel area.

The energy consumption consists of four main components:
\begin{equation}
E_{\text{consumption}}= E_{\text{drag}} + E_{\text{rolling}}+ E_{\text{gravitational}} + E_{\text{system}}.
\end{equation}
Key components are:
\begin{align}
&E_{\text{drag}}(r_i, t_i) = \frac{1}{2} C_d \rho A_{\text{frontal}} \left[v+v_{w}(r_i, t_i) \right]^2 d,\\
&E_{\text{rolling}}(r_i, t_i) = m g C_{\text{rr}} \cos( \theta_\text{r} )d,\\
&E_{\text{gravitational}}(r_i, t_i) = m g h, \\
&E_{\text{system}}(t_i) = P_{\text{loss}} t_i,
\end{align}
where $C_d$ is drag coefficient, $\rho$ is air density, $A_{\text{frontal}}$ is frontal area, $v$ is the velocity of the vehicle, $v_{w}(r_i, t_i)$ is the vehicle of the wind relative
to the heading of the vehicle, $d$ is distance during driving $m$ is vehicle mass, $g$ is gravitational acceleration, $C_{\text{rr}}$ is rolling resistance coefficient, $\theta_{\text{r}}$ is road incline, $h$ is elevation change, and $P_{\text{loss}}$ is constant power loss.

By expressing the energy components as functions of position, time, speed, and environmental variables, we developed a model that captures the dynamic interplay between vehicle speed and changing conditions. This model enables the calculation of energy balance for different speed profiles, supporting the development of optimized driving strategies.
\section{Experiments}

To validate our simulator's applicability for long-distance driving, we tested it using the World Solar Challenge (WSC) route, where solar vehicles travel approximately 3,020 km from Darwin to Adelaide within 8 days. This route is suitable for analyzing responsiveness to changes in solar irradiance and the impact of speed on long-distance driving.

We adhered to the WSC competition's driving conditions: charging is allowed only from 06:30 to 19:00 (operating time), calculated based on GHI data accounting for system losses, and driving is permitted from 08:00 to 17:00 (driving time).

We simulated the following strategies to understand solar vehicle operation:

\begin{itemize} 
\item \textbf{Plan 1 Minimum speed strategy}: Prioritizes stability but risks delays and inefficient energy use.
\item \textbf{Plan 2 Maximum speed strategy}: Maximizes speed but frequent recharging limits efficiency.
\item \textbf{Plan 3 Average speed strategy}: Maintains a steady speed, assuming ample energy availability.
\item \textbf{Plan 4 Daily average speed strategy}: Adjusts speed daily based on solar conditions for optimal energy use. 
\item \textbf{Plan 5: SoC maintenance strategy}: Balances speed and energy to keep the battery charge within an optimal range. \end{itemize}

Each strategy was simulated over the entire route using identical initial conditions and environmental data for comparison.

\begin{figure} [t]
\centering
\includegraphics[width=\columnwidth]{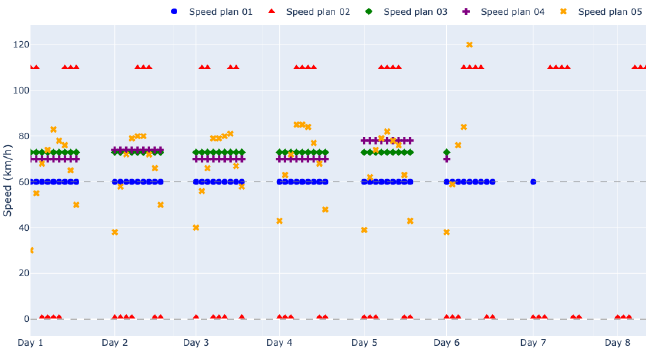}
\caption{Simulator result with SoC maintenance strategy scenario.}
\label{fig:plot_plans}
\end{figure}

\begin{table}
\caption{}
\centering
\begin{tabular}{l|cccccccc}
\hline
Plans & D1 & D2 & D3 & D4 & D5 & D6 & D7 & D8 \\
\hline
1 & 510 & 480 & 503 & 510 & 480 & 510 & 32 &  \\
2 & 499 & 281 & 430 & 390 & 390 & 390 & 342 & 303 \\
3 & 588 & 620 & 558 & 620 & 584 & 55 &  &  \\
4 & 588 & 622 & 560 & 595 & 624 & 36 &  &  \\
5 & 540 & 533 & 533 & 573 & 541 & 305 &  &  \\
\hline
\end{tabular}
\label{tab:daily-distances}
\end{table}

\section{Results}
Table \ref{tab:daily-distances} presents the daily driving distances for the five strategies. Under the same weather conditions and route, the Daily Average Speed Strategy resulted in the longest total distance covered. The second fastest arrival was achieved with the Average Speed Strategy. Driving at maximum speed with repeated charging and discharging led to the latest arrival. Although the Minimum Speed Strategy was slower, it resulted in a faster arrival than the maximum speed approach. Using the SoC Maintenance Strategy, the vehicle reached the destination over six days, arriving faster than with the minimum and maximum speed strategies. These results confirm that speed policies significantly impact solar vehicle driving, and drivers can establish and evaluate optimal strategies using the simulator. Figure \ref{fig:plot_plans} shows the hourly driving speeds for each strategy. In the Minimum Speed, Daily Average Speed, and Average Speed strategies, the vehicle's speed remains constant during the day's driving, allowing continuous travel without stops.

On Day 5, the daily average speed was higher than the overall route average speed due to a decrease in maximum environmental temperature from 29°C to 15°C when moving from Coober Pedy to Adelaide, increasing energy production efficiency by 2\% (Table \ref{tab:daily-weather-energy-data-beamsearch}). On other days, the speeds were similar to or lower than the overall route average speed. When driving at maximum speed, the maximum continuous driving time was 3 hours, varying with region and weather conditions. With the SoC Maintenance Strategy, no battery overload occurred, allowing efficient energy use. However, as it has a speed variance from 30km/h to 85km/h, it could cause driver fatigue with changing speed.

\begin{table}[t]

\caption{}
\centering
\begin{tabular}{l|cccccc}
\hline
Data & Day1 & Day2 & Day3 & Day4 & Day5 & Day6 \\
\hline
GHI & 5.476 & 5.220 & 4.897 & 4.826 & 4.389 & 2.210 \\
Temperature & 29.6 & 24.0 & 22.6 & 18.2 & 17.2 & 13.8 \\
Wind direction & SE & E & W & S & N & N \\
Wind speed & 5.0 & 5.4 & 3.3 & 3.5 & 2.1 & 4.6 \\
\hline
Drag & 2.450 & 2.985 & 2.784 & 2.994 & 2.600 & 1.445 \\
Rolling & 3.180 & 3.137 & 3.140 & 3.368 & 3.183 & 1.782 \\
Gravitational & 0.130 & 0.117 & 0.075 & -0.156 & -0.156 & 0.011 \\
Consumption & 6.372 & 6.815 & 6.575 & 6.798 & 6.215 & 3.557 \\
Generation & 7.013 & 7.427 & 7.279 & 7.390 & 6.677 & 3.166 \\
\hline
\end{tabular}%
\label{tab:daily-weather-energy-data-beamsearch}
\end{table}

\section{Conclusion}
By applying real driving routes and weather data, our simulator demonstrated how solar vehicles' driving range and speed policies change under various scenarios. This lets drivers predict energy states before driving and make appropriate speed decisions, facilitating thoughtful driving strategies. Additionally, it allows evaluation of the vehicle's design by changing the vehicle's specifications. While the simulator aims to provide results close to actual conditions, limitations exist, such as not incorporating real-time weather information and not reflecting urban shadow effects. Future research will enhance accuracy by integrating real-time data and additional spatial information.

Nevertheless, our simulator reliably imports data for various routes and credible weather information to calculate energy consumption similar to actual driving. This allows users to confirm solar vehicle driving results and deepen their understanding of energy management. Additionally, data generated by the simulator can serve as a foundation for advancing from traditional physical dynamics to AI, enabling developments like generating optimal speed policies for solar vehicles and creating dedicated support systems for solar driving.

\section*{Acknowledgment}
This work is supported by Korea Ministry of Land, Infrastructure and Transport (MOLIT) as 「Innovative Talent Education Program for Smart City」and supported by Korea Institute for Advancement of Technology (KIAT) grant funded by the Korea Government (MOTIE) (P0017304, Human Resource Development Program for Industrial Innovation)

\bibliographystyle{unsrtnat}
\bibliography{references}
\end{document}